\documentclass[letterpaper, 10 pt, conference]{ieeeconf}  

\IEEEoverridecommandlockouts                              

\usepackage{graphicx}
\usepackage{amsmath, amssymb}
\usepackage{graphics} 
\usepackage{epsfig} 
\usepackage{mathptmx} 
\usepackage{times} 
\usepackage{bm}
\usepackage{booktabs}
\usepackage{array}

\usepackage{color}
\usepackage{makecell}
\usepackage{multirow}
\usepackage{cite}
\usepackage{hyperref}

\title{\LARGE \bf
BoRe-Depth: Self-supervised Monocular Depth Estimation with Boundary Refinement for Embedded Systems
}

\author{Chang Liu$^{1}$, Juan Li$^{1*}$, Sheng Zhang$^{1}$, Chang Liu$^{2}$, Jie Li$^{1}$ and Xu Zhang$^{1}$
\thanks{This work was supported in part by the National Natural Science Foundation of China under Grant 62373053.}
\thanks{$^{*}$corresponding author: juanli@bit.edu.cn.}
\thanks{$^{1}$Chang Liu, Juan Li, Sheng Zhang, Jie Li and Xu Zhang are with School of Mechatronical Engineering, Beijing Institute of Technology, China.}
\thanks{$^{2}$Chang Liu is with Yangtze Delta Region Academy of Beijing Institute of Technology Jiaxing, China.}
}

\graphicspath{{Fig/}}

\begin{document}

\maketitle
\thispagestyle{empty}
\pagestyle{empty}

\begin{abstract}

Depth estimation is one of the key technologies for realizing 3D perception in unmanned systems. Monocular depth estimation has been widely researched because of its low-cost advantage, but the existing methods face the challenges of poor depth estimation performance and blurred object boundaries on embedded systems. In this paper, we propose a novel monocular depth estimation model, BoRe-Depth, which contains only 8.7M parameters. It can accurately estimate depth maps on embedded systems and significantly improves boundary quality. Firstly, we design an Enhanced Feature Adaptive Fusion Module (EFAF) which adaptively fuses depth features to enhance boundary detail representation. Secondly, we integrate semantic knowledge into the encoder to improve the object recognition and boundary perception capabilities. Finally, BoRe-Depth is deployed on NVIDIA Jetson Orin, and runs efficiently at 50.7 FPS. We demonstrate that the proposed model significantly outperforms previous lightweight models on multiple challenging datasets, and we provide detailed ablation studies for the proposed methods. The code is available at \href{https://github.com/liangxiansheng093/BoRe-Depth}{https://github.com/liangxiansheng093/BoRe-Depth}.

\end{abstract}

\section{INTRODUCTION}

Monocular depth estimation can quickly predict the dense depth maps from a single image. It is widely used in unmanned system navigation \cite{zheng2024monocular, kong2024robodepth}, autonomous driving\cite{shim2023swindepth, li2021unsupervised, guizilini2020semantically}, and augmented reality\cite{ganj2024hybriddepth,wu2022toward}. Existing studies\cite{liu2022lightweight,anantrasirichai2021fast,rudolph2022lightweight} have achieved real-time monocular depth estimation on embedded systems by designing lightweight models and optimizing network architectures. However, the depth maps generated by these models often lack details, appearing overly smooth and failing to provide accurate object boundaries. Such shortcomings may lead to object shape distortions and even fragmentation of a single object. As a result, models struggle to deliver the accurate and clear depth estimation results in practical applications.

\begin{figure}[t]  
    \centering
    \includegraphics[width=8cm]{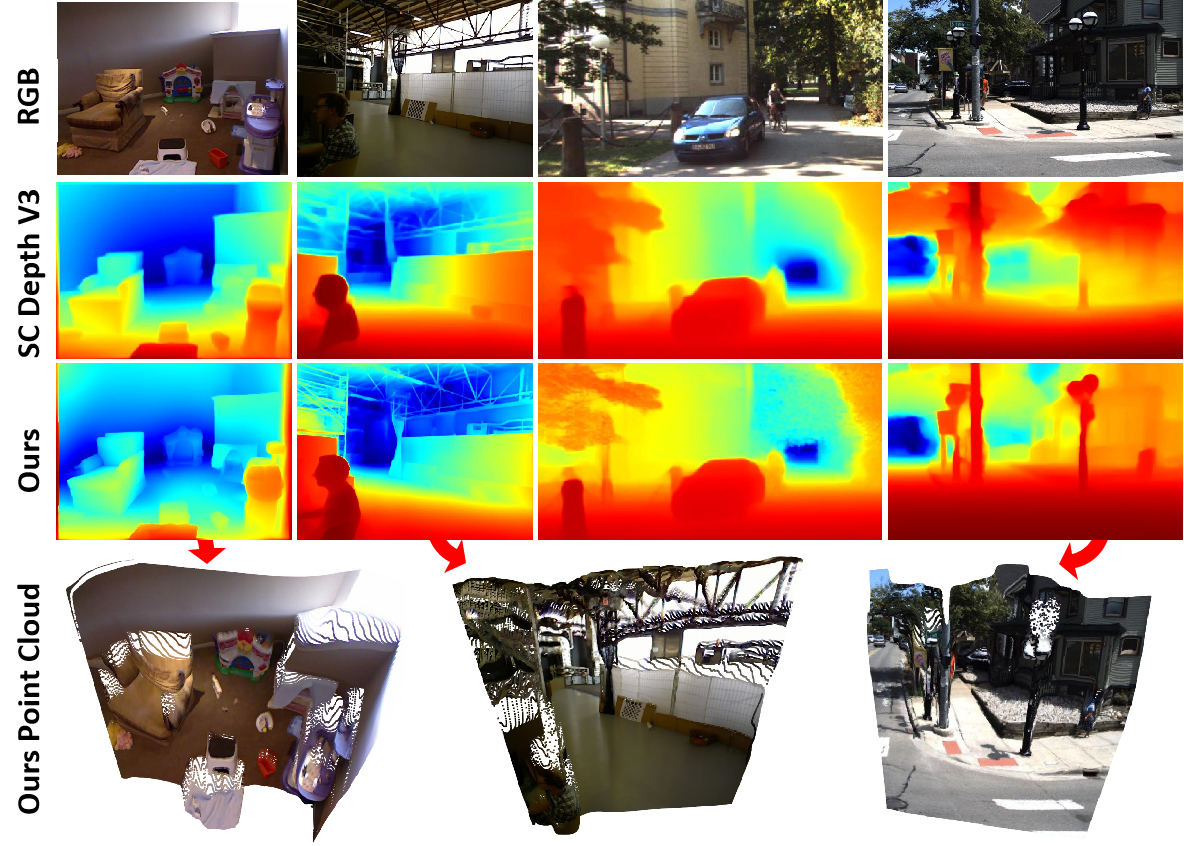}
    \caption{The proposed BoRe-Depth is a lightweight model with boundary refinement capability. It can refine more accurate boundary details and improve the quality of the dense point cloud.}
    \label{Fig.1}
\end{figure}

Currently, self-supervised monocular depth estimation models typically rely on view reconstruction loss\cite{zhou2017unsupervised} and geometric consistency loss\cite{bian2019unsupervised} during training. However, due to the small proportion of the boundary region, the losses in boundary areas receive insufficient attention. This neglect allows the model to generate seemingly high-quality depth estimation results, but the boundary pixels of the result are not captured accurately, leading to blurring of the generated depth map. Some studies have proposed methods\cite{bochkovskii2024depth,pham2024sharpdepth,ramamonjisoa2019sharpnet} to address this problem. However, these methods are difficult to meet the real-time computational requirements of embedded platforms because of a large number of parameters. Therefore, a more effective balance model in boundary refinement and real-time performance is urgently needed.

\begin{figure*}[thpb]  
    \centering
    \vspace{0.2cm} 
    \includegraphics[width=14.3cm]{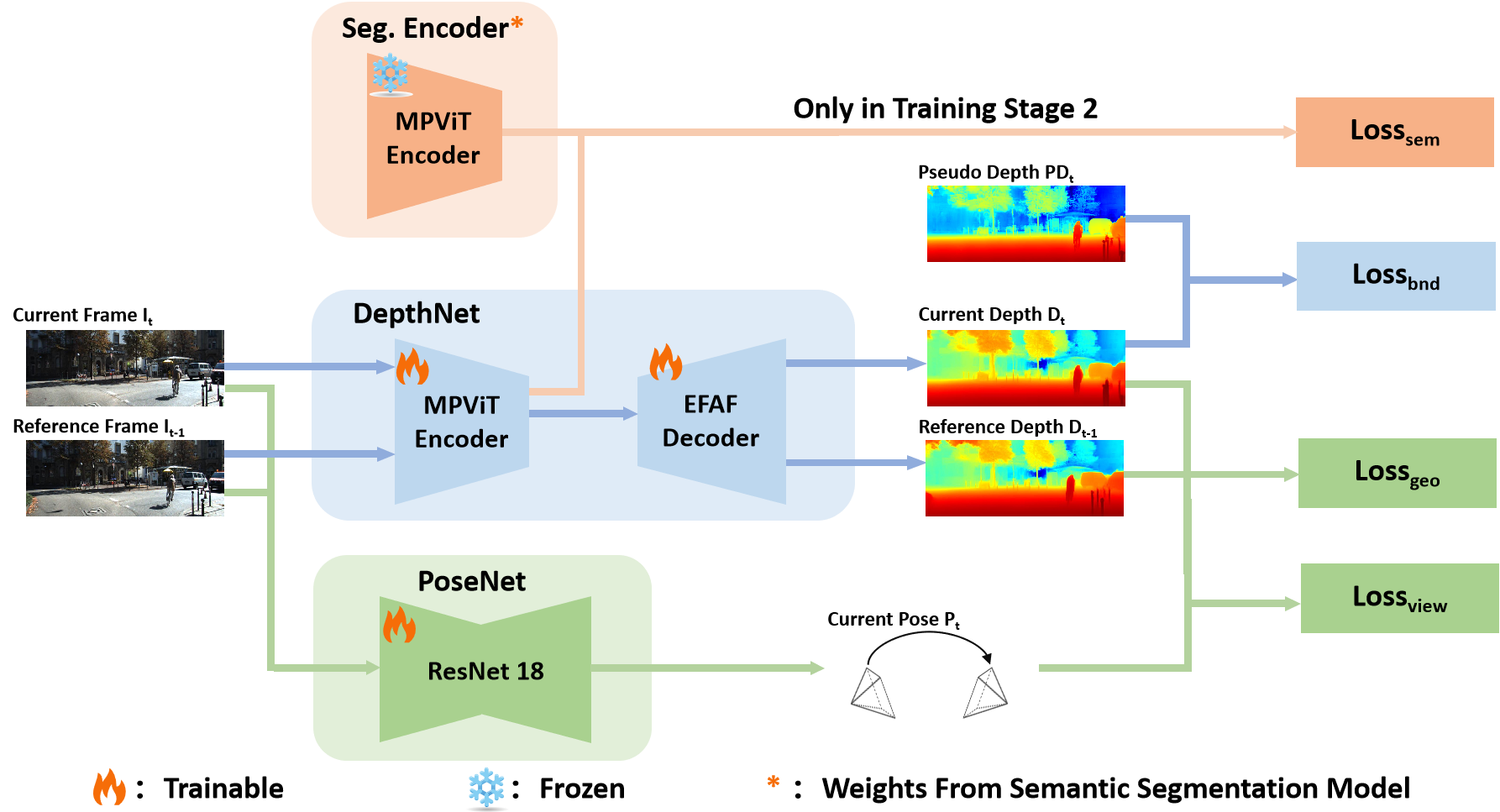}
    \caption{Overview of BoRe-Depth architecture. During training, the orange part represents $\textcolor[RGB]{247,177,135}{\textnormal{Semantic Segmentation Encoder}}$ introduced in the second stage, which calculates semantic information loss through the differences between features. The blue part represents $\textcolor[RGB]{189,215,238}{\textnormal{DepthNet}}$, which directly predicts the depth estimation result and calculates the boundary alignment loss through the pseudo-depth labels. The green part represents $\textcolor[RGB]{169,209,143}{\textnormal{PoseNet}}$, which computes the camera pose between two frames. It warps the images to calculate the geometric consistency loss and view reconstruction loss.}
    \label{Fig.2}
\end{figure*}

In this paper, we propose BoRe-Depth, a model that performs real-time high-quality depth estimation on embedded systems. We note that existing lightweight monocular depth estimation models\cite{yu2023udepth,poggi2022real} adopt simple and repetitive upsampling operations aimed at computational efficiency, which fails to make full use of the encoded features. Inspired by information fusion methods, such as weighted fusion\cite{mungoli2023adaptive} and stacking\cite{xia2021multi}, we design the Enhanced Feature Adaptive Fusion Module (EFAF). This module expands feature dimensions and adaptively integrates multi-level features to enhance the global detail representation capability of the model. To further improve the boundary quality, we design a two-stage training strategy. In the first stage, a coarse model is trained in the same way as previous models. In the second stage, the semantic information loss function is introduced to guide the model to focus on boundary regions and enhance the boundary quality. Specifically, we calculate the difference between features generated by the pre-trained semantic segmentation encoder\cite{lee2022mpvit} and those from the depth estimation encoder as the key loss function in the second stage. Under this constraint, the encoder learns semantic knowledge and develops object recognition capabilities.

In summary, the contributions of this paper are as follows:

\begin{itemize}

\item We propose BoRe-Depth for robust self-supervised learning of monocular depth with high boundary quality.
\item We design the Enhanced Feature Adaptive Fusion Module (EFAF), which improves boundary quality in the depth estimation results.
\item We design a two-stage training strategy. The semantic information loss is introduced in the second stage to encourage BoRe-Depth to learn semantic knowledge and capture object boundaries.
\item BoRe-Depth, with only 8.7M parameters, is optimized for embedded systems and runs at 50.7 FPS on NVIDIA Jetson Orin.

\end{itemize}

\section{RELATED WORK}

\subsection{Self-Supervised Monocular Depth Estimation}

 Massive amounts of ground-truth depth labels require a labor-intensive process for collection and cleaning. To overcome this problem, some researchers have innovatively proposed self-supervised monocular depth estimation models. These models achieve geometric view reconstruction from stereo image pairs\cite{garg2016unsupervised,godard2019digging} or sequential video frames\cite{xie2016deep3d,zhou2017unsupervised,bian2019unsupervised}, and compute losses based on geometric constraints to avoid the need of large-scale depth data. In current self-supervised models, various loss functions have been proposed to introduce additional constraints, such as left-right disparity consistency loss\cite{godard2017unsupervised}, photometric loss\cite{zhan2018unsupervised} and symmetry loss\cite{zhou2019unsupervised}. In recent years, some studies have proposed pseudo-depth\cite{sun2023sc, tan2023deep}. They are used for model training as more accessible data generated by large-scale depth estimation models.

\subsection{Lightweight Monocular Depth Estimation}

The parameter size and computational efficiency of monocular depth estimation models are crucial for deployment on embedded systems. Some studies\cite{wang2021knowledge,spek2018cream} explored how lightweight can be achieved through model compression, such as teacher-student networks and distillation learning. Other works\cite{wofk2019fastdepth} used network pruning and lightweight convolution structures to achieve model acceleration. These approaches are highly generalizable and can be combined with other techniques to further optimize lightweight models for embedded system applications. Additionally, some studies\cite{zheng2024monocular,sun2023sc} focused on designing small models to improve inference speed by reducing computational complexity. Recently, some scholars\cite{zhang2023lite,zhou2021r,yu2023udepth} proposed novel hybrid CNN-Transformer architectures, which successfully realized small-parameter models and achieved significant results.

\subsection{Boundary-Refined Monocular Depth Estimation}

Accurately delineating object boundaries remains a significant challenge in monocular depth estimation. Some researchers\cite{ramamonjisoa2019sharpnet,xue2021boundary} introduced manually annotated boundary datasets during training, and used the boundary information as prior knowledge to constrain model learning. However, it is still a difficult task to manually annotate fine object boundaries. Some studies\cite{jung2021fine,ochs2019sdnet,miclea2023dynamic} incorporated semantic segmentation into depth estimation. They guided depth estimation models to recognize objects and enhance their focus on object boundaries in the scene. In addition, some scholars\cite{bochkovskii2024depth,lyu2021hr} obtained multi-level features through repeated downsampling, and then fused them in a coarse-to-fine method to obtain depth estimation results with refined boundaries. In recent years, some studies have introduced the diffusion model to enhance the boundary details in depth estimation. For instance, Marigold\cite{ke2024repurposing}, DepthFM\cite{gui2024depthfm}, and other works\cite{saxena2024surprising} leverage multi-step generation and denoising processes to finely control boundary details.

\section{METHODOLOGY}

\subsection{Overview}

Our goal is to design a real-time monocular depth estimation model with high-quality boundaries. Through the joint constraint of pseudo-depth labels and the semantic segmentation encoder, our model achieves both excellent depth estimation accuracy and boundary quality. Fig.\ref{Fig.2} illustrates an overview of BoRe-Depth. We design lightweight EFAF and incorporate it into the EFAF Decoder of DepthNet. Additionally, we design a two-stage training strategy and introduce the semantic information loss function in the second stage to further improve the boundary quality.

\subsubsection{Pseudo Depth Labels} 

During training, we use a large monocular depth estimation model to obtain pseudo-depth labels, which have clearer boundaries than the ground-truth depth labels. These labels provide the excellent boundary references to encourage our method to effectively capture object boundaries. However, although pseudo-depth labels have obvious advantages of easy access and high boundary quality, their accuracy is inevitably limited by the inherent errors of the model itself. Therefore, they cannot completely replace ground-truth depth labels. During validating, we no longer use pseudo-depth labels but use ground-truth depth labels to evaluate the depth estimation accuracy. This ensures that the model achieves reliable depth prediction in practice while effectively correcting the potential biases introduced by pseudo-depth labels.

\subsubsection{Encoder}

Exploring global information is crucial for monocular depth estimation, so the backbone network with strong feature representation capability is required to infer the contextual information. In this paper, we choose MPViT architecture\cite{lee2022mpvit} as the encoder. In previous studies, the effectiveness of MPViT has been verified in monocular depth estimation tasks, such as GasMono\cite{zhao2023gasmono}. It consists of a stem layer and four transformer encoders. For a given input image $I \in \mathbb{R}^{H \times W \times 3}$, five layers of depth features $F^{i},i=1,2,3,4,5$ can be generated. By leveraging the multi-path mechanism, MPViT independently feeds tokens of different scales into multiple transformer encoders, and aggregates the generated multi-level features, thereby achieving both fine-grained and coarse-grained depth feature representation. Furthermore, the lightweight design of MPViT is another reason for our choice. For example, the MPViT-tiny model has only 5.8M parameters, greatly reducing computational costs while maintaining efficient feature encoding.

Regarding the decoder, we carefully designed EFAF to improve the boundary quality, which will be detailed in \ref{subsec:EFAF}. The semantic information loss function will be described in detail in \ref{subsec:Semantic Information Loss Function}. Finally, the two-stage training strategy will be described in \ref{subsec:Training Strategy}.

\begin{figure}[thpb]  
    \centering
    \vspace{0.2cm} 
    \includegraphics[width=8.1cm]{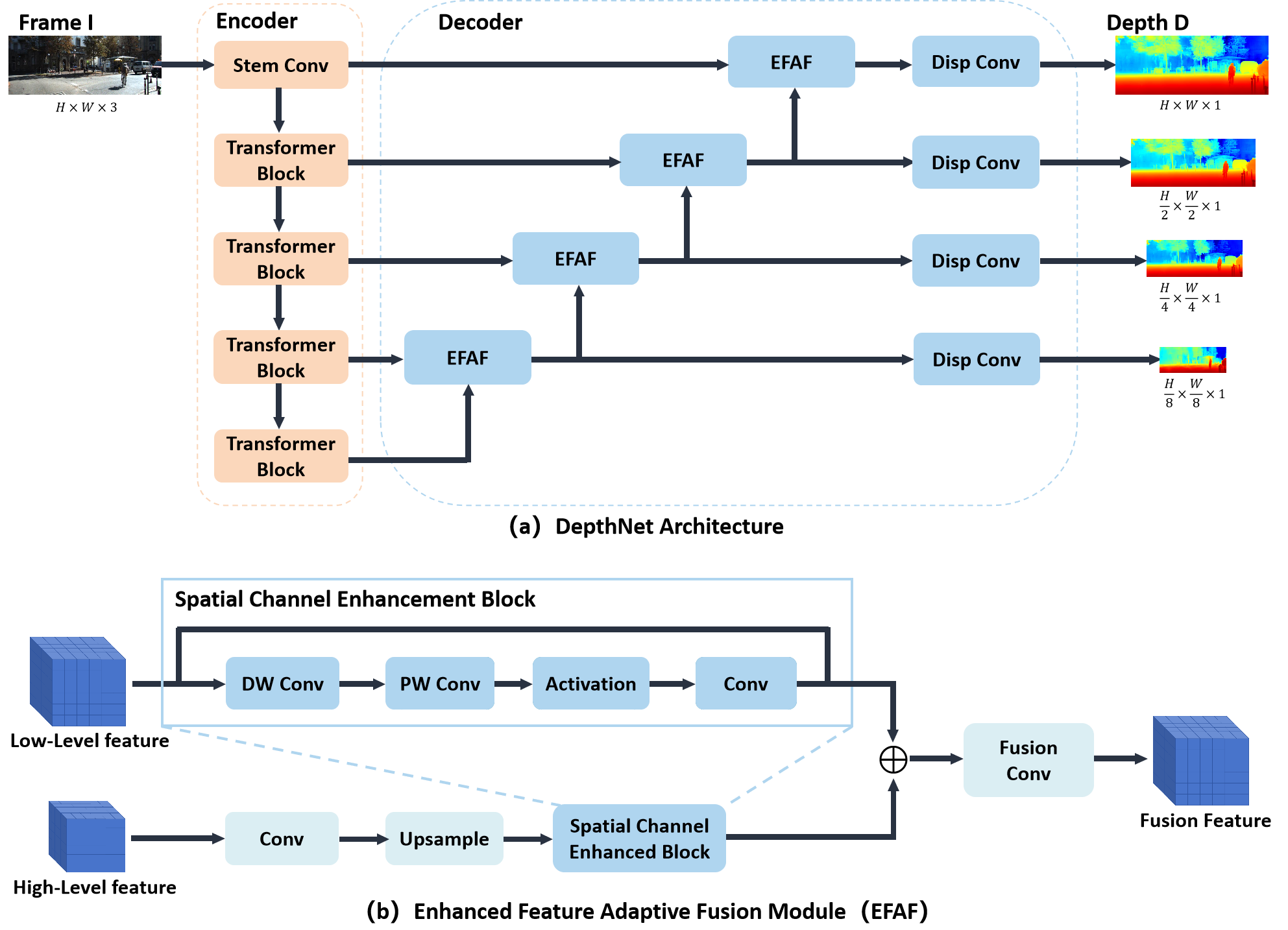}
    \caption{DepthNet network architecture. (a) The overall architecture of the depth estimation network is presented. This network effectively extracts multi-scale features through the encoder-decoder structure and generates high-quality depth maps. (b) The EFAF module is demonstrated, which aggregates features at each level through lightweight convolution, thereby improving the boundary quality.}
    \label{Fig.3}
\end{figure}

\subsection{EFAF Decoder}
\label{subsec:EFAF}

\subsubsection{Decoder}

The entire decoder design follows the hierarchical feature aggregation strategy, progressively refining the depth estimation results from coarse to fine. As illustrated in Fig.3(a), the decoder receives depth features from five layers of the MPViT encoder and aggregates the information layer by layer to achieve fine depth prediction. Features of adjacent layers are first channel-enhanced by EFAF to enrich the boundary details across the global image. This hierarchical network takes full advantage of the complementarity between different resolution features, allowing the decoder to capture spatial structures and depth variations more accurately in complex scenes. At the same time, the strategy of gradual size recovery reduces the feature losses and helps the decoder to generate higher-quality depth estimation results from coarse to fine.

\subsubsection{Enhanced Feature Adaptive Fusion Module}

High-quality boundary details are essential for generating clear depth estimation results. To better aggregate depth features, we design an Enhanced Feature Adaptive Fusion module(EFAF) based on the original feature decoder. EFAF enhances key features by adaptively fusing adjacent features, so as to improve the ability of the decoder to capture object boundaries. The specific architecture of EFAF is shown in Fig.\ref{Fig.3}(b). One of the key components is the Spatial Channel Enhancement Block(SCE). This module bolsters object boundary details by expanding feature dimensions and incorporates lightweight convolution,such as depth-wise convolution (DW Conv) and point-wise convolution (PW Conv)\cite{howard2017mobilenets}, to reduce the computational cost. More precisely, SCE first generates high-dimensional implicit features to extend the spatial representation of depth features. Then they are sequentially fed into the channel-adaptive convolution to obtain the channel-adaptive features $F_{d}^{i}$, and then into the skip connection to obtain the features after channel enhancement $F_{ce}^{i}$. These steps can be implemented as follows:

$$ 
F_{d}^{i} = \phi (PWConv_{1 \times 1}(DWConv_{3 \times 3}(F^{i}))),
\eqno{(1)}
$$

$$
F_{ce}^{i} = F^{i} + Conv_{1 \times 1}(F_{d}^{i}),
\eqno{(2)}
$$
where $\phi(\cdot)$ represents the activation function GELU, and "+" means the skip connection.

It should be noted that features at different levels do not share weights when they pass through SCE, because the content of feature representation is different in the feature adaptive enhancement process. The feature aggregation is carried out through concatenation and fusion convolution(Fusion Conv) after two features are independently enhanced. Formally,

$$
F_{fusion}^{i} = 
    \begin{cases}
        Conv(F_{ce}^{i} \oplus F_{fusion}^{i+1}),i=1,2,3 \\
        Conv(F_{ce}^{i} \oplus F_{ce}^{i+1}),i=4
    \end{cases},
\eqno{(3)}
$$
where $F_{fusion}^{i}$ means the detailed depth features after adaptive fusion, and $\oplus$ represents the concatenation operation.

\subsection{Semantic Information Loss Function}
\label{subsec:Semantic Information Loss Function}

In the past, many studies\cite{jung2021fine,ochs2019sdnet,miclea2023dynamic} have attempted to improve depth estimation performance by incorporating semantic segmentation. We also believe that fusing different pixel-level scene perception information can effectively improve depth estimation results, particularly boundary quality. Based on this\cite{ochs2019sdnet}, we design a shared MPViT encoder and two independent decoders to jointly train monocular depth estimation and semantic segmentation. However, as shown in Table \ref{table.4}, the experimental results indicate that the gains from this approach are limited. We believe that the root cause of this limitation is that when depth estimation is the main task, the presence of the semantic segmentation decoder causes large changes in its internal parameter weights during optimization. The semantic information obtained from the joint task by the shared encoder does not have a significant effect.

To address this impact, we introduce two independent MPViT encoders, one for extracting semantic segmentation features and the other for extracting depth estimation features respectively, and compare the feature similarity between them. Firstly, the MPViT encoder is trained on semantic segmentation tasks to obtain semantic knowledge. The encoder is then frozen and used as a branch network to guide the depth estimation encoder to learn semantic information. In this way, our lightweight depth estimation model benefits from semantic knowledge while avoiding the potential negative effects of semantic segmentation decoders. In terms of implementation, we design a contrastive constraint on the pixel-level features generated by two independent encoders. The semantic segmentation feature guides the depth estimation encoder to acquire semantic knowledge by calculating the similarity between semantic segmentation features and monocular depth estimation features. The specific formula of the semantic information loss function $L_{sem}$ is as follows:

$$
L_{sem} = 1 - \frac {1} {N} \sum_{i=1}^{N} \frac {F^{i} \cdot F_{ss}^{i}} {\vert\vert {F^{i} \vert\vert_{2} \cdot \vert\vert F_{ss}^{i}} \vert\vert_{2}},
\eqno{(4)}
$$
where $N=5$ represents the total number of feature layers and $F_{ss}^{i}$ represents the semantic segmentation features of the $i^{th}$ layer.

\subsection{Training Strategy}
\label{subsec:Training Strategy}

To maximize the potential of the semantic information loss function, we design a two-stage training strategy. Unlike previous studies\cite{yang2024depth} that directly introduce semantic and other losses in the single stage, our two-stage strategy progressively improves the model under the constraints of different loss combinations.

\subsubsection{The First Stage}

The first is view reconstruction loss. For two adjacent frames $I_{t-1}$ and $I_{t}$, we predict the 6D camera pose $P_{t}$ through PoseNet. Then $I_{t-1}$ synthesizes $I_{t}^{'}$ using the warping flow. The view reconstruction loss $L_{view}$ is computed using the Structural Similarity Index(SSIM)\cite{wang2004image} and L1 regularization loss. Formally,

$$
L_{view} = (1- \lambda) \vert\vert I_{t}-I_{t}^{'}\vert\vert_{1} + \lambda \frac{1-SSIM(I_{t},I_{t}^{'})}{2}.
\eqno{(5)}
$$
In general, $\lambda$ is set to 0.85.

Next is the geometric consistency loss\cite{bian2019unsupervised}. We encourage the depth images to maintain reasonable geometric variation between adjacent frames. The geometric consistency loss $L_{geo}$ is calculated as follows:

$$
L_{geo}=Diff_{geo}(\widehat{D}_{t}, D_{t}^{'}),
\eqno{(6)}
$$
where $Diff_{geo}(\cdot)$ represents the loss between the predicted depth map and the warped depth map, $\widehat{D}_{t}$ denotes the depth map predicted by the model, and $D_{t}^{'}$ represents the depth map synthesized from $D_{t-1}^{'}$ through the warping flow.

Finally, we calculate the boundary alignment loss $L_{bnd}$ using pseudo-depth labels. The shape and structure of objects are jointly constrained by the normal and boundary. Formally,

\begin{figure*}  
    \centering
    \vspace{0.2cm} 
    \includegraphics[width=13.5cm]{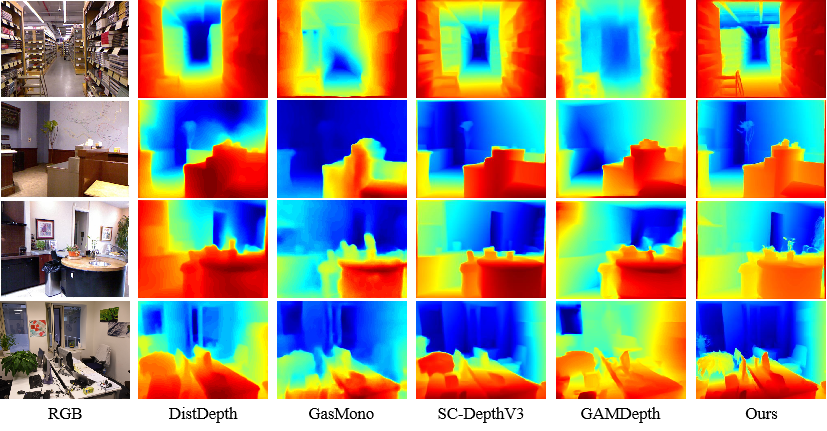}
    \caption{Qualitative indoor depth estimation results. Four images are respectively from NYUv2 dataset and IBims-1 dataset. Existing models are hard to describe the object boundaries quickly, which leads to blurred depth estimation. In contrast, our model predicts the most accurate depth with the clearest boundaries robustly.}
    \label{Fig.4}
\end{figure*}

\begin{figure*}  
    \centering
    \vspace{0.2cm} 
    \includegraphics[width=13.5cm]{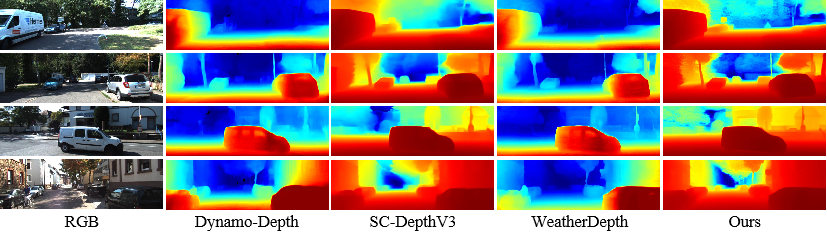}
    \caption{Qualitative outdoor depth estimation results. Four images are from KITTI dataset. Obviously, our model has the best estimation accuracy as well as boundary quality.}
    \label{Fig.5}
\end{figure*}

$$
L_{bnd} = \theta \cdot f(\nabla \widehat{D}_{t}, \nabla PD_{t}) + \vartheta \cdot f(\partial \widehat{D}_{t}, \partial PD_{t}),
\eqno{(7)}
$$
where the values of $\theta$ and $\vartheta$ are typically equal and set to 0.1, $f(\cdot)$ represents the similarity function, $\nabla$ and $\partial$ denote the normal and boundary of the depth map respectively. The normal is calculated by the gradient, and the boundary is calculated by the 3×3 Sobel operator.

It is important to note that ground-truth depth labels inevitably contain invalid zero points. When using the ground-truth depth labels during validation, all loss function calculations are performed only at valid points.

In the first stage, our loss function $L_{total}^{stage1}$ can be expressed as:

$$
L_{total}^{stage1}= \alpha L_{view} + \beta L_{geo} + \gamma L_{bnd}.
\eqno{(8)}
$$
Based on previous experience, when $\alpha=1$, $\beta=1$ and $\gamma=1$ are set in the formula, the model will perform better in the first stage.

\subsubsection{The Second Stage}

In the second stage, our goal is for the model to better understand the semantic information to describe the object boundaries in the image. We introduce the semantic information loss to optimize depth features generated by the encoder. The purpose is to avoid potential issues that may arise in the first stage, that is, the semantic information loss is miscalculated because of the pretrained classification encoder, leading to an incorrect local optimum. As described in \ref{subsec:Semantic Information Loss Function}, we constrain the differences between the features of the semantic segmentation encoder and those of the monocular depth estimation encoder by comparing their similarity. The core idea of this difference constraint is that the features generated by the depth estimation encoder should be similar to those of the semantic segmentation encoder, so that the knowledge of the semantic segmentation encoder can be transferred effectively. The loss function of the second stage $L_{total}^{stage2}$ can be expressed as:

\begin{table*}[ht]
\vspace{0.2cm} 
\caption{Self-supervised monocular depth estimation results.  $\bf{Bold}$ indicates the best and \underline{underlined} indicates the second. $\downarrow$ indicates that the lower the effect is better, and $\uparrow$ indicates that the higher the better.}
\label{table.1}
\begin{center}
\begin{tabular}{c|c|c|c|c|c|c|c|c}
\hline
Dataset & Models & Param./M$\downarrow$ & Abs.Rel$\downarrow$ & RMSE $\downarrow$ & $\delta_{1} \uparrow$ & $\delta_{2} \uparrow$ & $\delta_{3} \uparrow$ & $\epsilon_{DBE}^{acc} \downarrow$\\
\hline
\multirow{5}*{NYUv2} & DistDepth\cite{wu2022toward} & 69.2 & $\underline{0.113}$ & $\underline{0.444}$ & $\underline{0.873}$ & $\bm{0.974}$ & $\bm{0.993}$ & 3.359\\
~ & GasMono\cite{zhao2023gasmono} & 28.0 & $\underline{0.113}$ & 0.459 & 0.871 & $\underline{0.973}$ & $\underline{0.992}$ & 3.463\\
~ & SC-DepthV3\cite{sun2023sc} & 18.4 & 0.123 & 0.486 & 0.848 & 0.963 & 0.991 & $\underline{2.435}$\\
~ & GAM-Depth\cite{cheng2024gam} & $\underline{14.8}$ & 0.131 & 0.507 & 0.836 & 0.960 & 0.990 & 4.325\\
\cline{2-9}
~ & BoRe-Depth(Ours) & $\bm{8.7}$ & $\bm{0.101}$ & $\bm{0.429}$ & $\bm{0.883}$ & 0.971 & $\bm{0.993}$ & $\bm{2.083}$\\
\Xhline{1px}
\multirow{5}*{KITTI} & Lite-Mono\cite{zhang2023lite} & $\bm{3.1}$ & 0.107 & 4.561 & 0.886 & 0.963 & 0.983 & 3.357\\
~ & Dynamo-Depth\cite{sun2024dynamo} & 46.2 & 0.112 & 4.505 & 0.873 & 0.959 & $\underline{0.984}$ & 3.799\\
~ & SC-DepthV3\cite{sun2023sc} & 18.4 & 0.118 & 4.709 & 0.864 & 0.960 & $\underline{0.984}$ & $\underline{2.780}$\\
~ & WeatherDepth\cite{wang2024weatherdepth} & 27.9 & $\underline{0.104}$ & $\underline{4.483}$ & $\underline{0.887}$ & $\underline{0.965}$ & $\underline{0.984}$ & 3.119\\
\cline{2-9}
~ & BoRe-Depth(Ours) & $\underline{8.7}$ & $\bm{0.103}$ & $\bm{4.323}$ & $\bm{0.889}$ & $\bm{0.967}$ & $\bm{0.986}$ & $\bm{2.649}$\\
\hline
\end{tabular}
\end{center}
\end{table*}

\begin{table*}
\caption{Zero-shot monocular depth estimation results on iBims-1 dataset. All models are trained on NYUv2 dataset only.}
\label{table.2}
\begin{center}
\begin{tabular}{c|c|c|c|c|c|c|c}
\hline
Models & Param./M$\downarrow$ & Abs.Rel$\downarrow$ & RMSE $\downarrow$ & $\delta_{1} \uparrow$ & $\delta_{2} \uparrow$ & $\delta_{3} \uparrow$ & $\epsilon_{DBE}^{acc} \downarrow$\\
\hline
DistDepth\cite{wu2022toward} & 69.2 & 0.226 & 1.054 & 0.770 & 0.889 & 0.949 & 4.883\\
GasMono\cite{zhao2023gasmono} & 28.0 & 0.252 & 0.809 & 0.699 & 0.853 & 0.932 & 3.441\\
SC-DepthV3\cite{sun2023sc} & 18.4 & $\underline{0.172}$ & $\underline{0.737}$ & $\underline{0.812}$ & $\underline{0.945}$ & $\bm{0.983}$ & $\underline{3.001}$\\
GAM-Depth\cite{cheng2024gam} & $\underline{14.8}$ & 0.197 & 0.968 & 0.800 & 0.919 & 0.968 & 5.849\\
\hline
BoRe-Depth(Ours) & $\bm{8.7}$ & $\bm{0.124}$ & $\bm{0.675}$ & $\bm{0.843}$ & $\bm{0.954}$ & $\bm{0.983}$ & $\bm{2.486}$\\
\hline
\end{tabular}
\end{center}
\end{table*}

$$
L_{total}^{stage2} = \varepsilon L_{sem} + L_{total}^{stage1},
\eqno{(9)}
$$
where $\varepsilon$ is a hyperparameter that balances the weight between semantic information loss and other losses. After testing, when $\varepsilon=0.01$, the embedding effect of semantic knowledge is the best. In this way, we achieve the optimal embedding of semantic information in the monocular depth estimation model, so that the model can not only pay attention to the geometric information of the scene, but also focus on the semantic information of the object in complex scenes. The model can significantly refine object details and reduce the errors caused by the blur of object boundaries.

\section{EXPERIMENTS}

\subsection{Implementation Details}

\subsubsection{Datasets}

Our method can be widely applied to monocular depth estimation tasks in various scenes. To validate its effectiveness, we selected two benchmark datasets, NYUv2\cite{silberman2012indoor} dataset for static indoor scenes and KITTI\cite{geiger2013vision} dataset for dynamic outdoor scenes. Additionally, the model trained on NYUv2 dataset is evaluated on IBims-1\cite{koch2018evaluation} dataset to evaluate the generalization capability.

\subsubsection{Training Details}

We implement BoRe-Depth using the PyTorch library. The model is trained for 100 epochs for each task, and the learning rate is $10^{-4}$. The weights of the optimal performance period during training are taken as the final result.

\subsubsection{Evaluation Metrics}

We adopt the standard evaluation metrics\cite{eigen2014depth} in monocular depth estimation, including absolute relative error(Abs\_Rel), root mean squared error(RMSE), and the accuracy under threshold($\delta_{1},\delta_{2},\delta_{3}$). 

Next, to clearly demonstrate the advantage of BoRe-Depth in terms of boundary quality, we introduced the accuracy metric error in the depth boundary error $\epsilon_{DBE}^{acc}$\cite{koch2018evaluation} to evaluate the boundary quality: 

$$
\epsilon_{DBE}^{acc}(\bm{Y})=\frac{1}{\sum_{i}{\sum_{j}{y_{bin;i,j}}}}\sum_{i}{\sum_{j}{e_{i,j}^{*} \cdot y_{bin;i,j}}},
\eqno{(10)}
$$
where $\bm{Y}$ represents the predicted depth map, $y_{bin;i,j} \in \bm{Y}_{bin}$ is the object boundary extracted using the structured edges, and $\bm{Y}_{bin}^{*}$ is the ground-truth boundary labels. The ground-truth boundary image $\bm{E}^{*}=DT(\bm{Y}_{bin}^{*})$ is generated through the Euclidean distance transform, where $e_{i,j}^{*}$ represents each pixel in $\bm{E}^{*}$.

Finally, we calculate the number of parameters for each model, which is closely related to the calculation speed and the occupied memory when the model is deployed.

\subsection{Evaluation Results}

We use datasets from different scenes above to evaluate the proposed model. Quantitative depth estimation results are shown in Table \ref{table.1}, while visual depth estimation results are shown in Fig.\ref{Fig.4} and Fig.\ref{Fig.5}. Next, these results will be analyzed in more detail.

\subsubsection{Results on Depth Accuracy and Object Boundary}

Table \ref{table.1} shows the experimental results on NYUv2 and KITTI datasets. The results demonstrate that BoRe-Depth achieves the state-of-the-art performance with the fewest parameters, particularly in the boundary quality. Notably, GasMono uses MPViT-small encoder which is a similar backbone to ours. Despite having a much larger network structure, it is still markedly inferior than BoRe-Depth especially in boundary quality.

It is worth mentioning that, through analyzing the experimental results, we believe that the improvement of boundary quality has a promoting effect on the accuracy of monocular depth estimation. The clear boundary enhances the detailed expression of complex structural regions. It helps the monocular depth estimation task to better recognize objects, thus improving the accuracy of results.

\subsubsection{Results on Zero-Shot Generalization}

We conduct the zero-shot monocular depth estimation testing on iBims-1 dataset to verify the generalization capability under new scenes. Models are trained only on NYUv2 dataset and evaluated on iBims-1 dataset. As shown in Table \ref{table.2}, the experimental results demonstrate that our model performs excellently in zero-shot generalization and can robustly handle monocular depth estimation tasks in new scenes.

\subsection{Ablation Studies}

\begin{table}
\vspace{0.2cm} 
\caption{Ablation studies of the proposed EFAF on NYUv2 dataset. "wo" is the abbreviation for "without".}
\label{table.3}
\begin{center}
\resizebox{0.48\textwidth}{!}{
\begin{tabular}{>{\centering\arraybackslash}m{2.3cm}|c|c|c|c|c}
\hline
Models & Param./M$\downarrow$ & Abs.Rel$\downarrow$ & RMSE $\downarrow$ & $\delta_{1} \uparrow$ & $\epsilon_{DBE}^{acc} \downarrow$\\
\hline
Baseline & $\bm{7.3}$ & 0.113 & 0.454 & 0.865 & 2.226\\
Baseline+EFAF wo high-level SCE & 8.1 & $\underline{0.108}$ & $\underline{0.450}$ & $\underline{0.872}$ & $\underline{2.158}$\\
Baseline+EFAF wo low-level SCE & $\underline{7.8}$ & 0.109 & 0.458 & 0.871 & 2.178\\
Baseline+EFAF & 8.7 & $\bm{0.106}$ & $\bm{0.439}$ & $\bm{0.874}$ & $\bm{2.129}$\\
\hline
\end{tabular}
}
\end{center}
\end{table}

\subsubsection{Enhanced Feature Adaptive Fusion Module}

We demonstrate the effectiveness of EFAF by removing some of SCE branches and evaluating them on NYUv2 dataset. Specific experimental results are shown in Table \ref{table.3}. The experimental results clearly demonstrate that each SCE branch improves the boundary quality in our model. During the feature fusion process, the features from both branches are combined with adaptive weights to enhance the boundary perception capability. Applying feature expansion to only one branch would cause the loss of key information from the other branch, leading to performance degradation. This experiment further verifies the necessity and effectiveness of each SCE block in EFAF.

\begin{table}[h]
\caption{Evaluation results of semantic information loss on NYUv2 dataset.}
\label{table.4}
\begin{center}
\resizebox{0.48\textwidth}{!}{
\begin{tabular}{>{\centering\arraybackslash}m{2cm}|c|c|c|c}
\hline
Models & Abs.Rel$\downarrow$ & RMSE $\downarrow$ & $\delta_{1} \uparrow$ & $\epsilon_{DBE}^{acc} \downarrow$\\
\hline
Baseline & 0.106 & 0.439 & 0.874 & 2.129\\
Baseline+semantic decoder & $\underline{0.104}$ & 0.434 & 0.877 & 2.122\\
Baseline+$L_{sem}$ in Stage1 & $\underline{0.104}$ & $\underline{0.433}$ & $\underline{0.878}$ & $\underline{2.107}$\\
Baseline+$L_{sem}$ in Stage2 & $\bm{0.101}$ & $\bm{0.429}$ & $\bm{0.883}$ & $\bm{2.083}$\\
\hline
\end{tabular}
}
\end{center}
\end{table}

\subsubsection{Semantic Information Loss and Training Strategy}

We also conducted the experiments to evaluate the effectiveness of the two-stage training strategy. By assessing our model trained with various strategies on NYUv2 dataset, as shown in Table \ref{table.4}, the experimental results clearly show that our strategy of setting semantic information loss in the second training stage prompts the model to achieve the best performance.

We believe that introducing semantic information is an effective strategy to improve monocular depth estimation accuracy. If semantic information is introduced too early in the first stage, the model may be affected due to the differences between classification features learned during pretraining and the introduced semantic segmentation features. Instead, we adopt a two-stage training strategy, where semantic information is introduced after the encoder has learned the coarse depth features. Introducing semantic information in the second stage can provide rich object shape information. In this way, the model gradually refines from the coarse depth estimation results. Because it can help the model to better understand the object structure and spatial relationship, so as to improve the performance of the model.

\section{CONCLUSIONS}

In this work, we propose BoRe-Depth, a method capable of real-time monocular depth estimation with high boundary quality on embedded systems. Our approach leverages a large depth estimation model to generate high-quality depth maps which serve as pseudo-depth labels for self-supervised learning. Regarding our design, we significantly improve monocular depth estimation accuracy and boundary quality through the carefully designed Enhanced Feature Adaptive Fusion Module (EFAF) and the semantic information loss function in the second stage. We validate our method on three datasets that cover indoor and outdoor, static and dynamic scenes. Experimental results demonstrate that our model remarkably outperforms various existing lightweight monocular depth estimation models. Further ablation experiments also validate the effectiveness of our proposed method, highlighting its crucial role in enhancing model performance.











\bibliographystyle{IEEEtran}

\bibliography{bibfile}

\begin{thebibliography}{10}
\providecommand{\url}[1]{#1}
\csname url@samestyle\endcsname
\providecommand{\newblock}{\relax}
\providecommand{\bibinfo}[2]{#2}
\providecommand{\BIBentrySTDinterwordspacing}{\spaceskip=0pt\relax}
\providecommand{\BIBentryALTinterwordstretchfactor}{4}
\providecommand{\BIBentryALTinterwordspacing}{\spaceskip=\fontdimen2\font plus
\BIBentryALTinterwordstretchfactor\fontdimen3\font minus \fontdimen4\font\relax}
\providecommand{\BIBforeignlanguage}[2]{{%
\expandafter\ifx\csname l@#1\endcsname\relax
\typeout{** WARNING: IEEEtran.bst: No hyphenation pattern has been}%
\typeout{** loaded for the language `#1'. Using the pattern for}%
\typeout{** the default language instead.}%
\else
\language=\csname l@#1\endcsname
\fi
#2}}
\providecommand{\BIBdecl}{\relax}
\BIBdecl

\bibitem{zheng2024monocular}
H.~Zheng, S.~Rajadnya, and A.~Zakhor, ``Monocular depth estimation for drone obstacle avoidance in indoor environments,'' in \emph{2024 IEEE/RSJ International Conference on Intelligent Robots and Systems (IROS)}.\hskip 1em plus 0.5em minus 0.4em\relax IEEE, 2024, pp. 10\,027--10\,034.

\bibitem{kong2024robodepth}
L.~Kong, S.~Xie, H.~Hu, L.~X. Ng, B.~Cottereau, and W.~T. Ooi, ``Robodepth: Robust out-of-distribution depth estimation under corruptions,'' \emph{Advances in Neural Information Processing Systems}, vol.~36, 2024.

\bibitem{shim2023swindepth}
D.~Shim and H.~J. Kim, ``Swindepth: Unsupervised depth estimation using monocular sequences via swin transformer and densely cascaded network,'' in \emph{2023 IEEE International Conference on Robotics and Automation (ICRA)}.\hskip 1em plus 0.5em minus 0.4em\relax IEEE, 2023, pp. 4983--4990.

\bibitem{li2021unsupervised}
H.~Li, A.~Gordon, H.~Zhao, V.~Casser, and A.~Angelova, ``Unsupervised monocular depth learning in dynamic scenes,'' in \emph{Conference on Robot Learning}.\hskip 1em plus 0.5em minus 0.4em\relax PMLR, 2021, pp. 1908--1917.

\bibitem{guizilini2020semantically}
V.~Guizilini, R.~Hou, J.~Li, R.~Ambrus, and A.~Gaidon, ``Semantically-guided representation learning for self-supervised monocular depth,'' \emph{arXiv preprint arXiv:2002.12319}, 2020.

\bibitem{ganj2024hybriddepth}
A.~Ganj, H.~Su, and T.~Guo, ``Hybriddepth: Robust depth fusion for mobile ar by leveraging depth from focus and single-image priors,'' \emph{arXiv e-prints}, pp. arXiv--2407, 2024.

\bibitem{wu2022toward}
C.-Y. Wu, J.~Wang, M.~Hall, U.~Neumann, and S.~Su, ``Toward practical monocular indoor depth estimation,'' in \emph{Proceedings of the IEEE/CVF conference on computer vision and pattern recognition}, 2022, pp. 3814--3824.

\bibitem{liu2022lightweight}
S.~Liu, L.~T. Yang, X.~Tu, R.~Li, and C.~Xu, ``Lightweight monocular depth estimation on edge devices,'' \emph{IEEE Internet of Things Journal}, vol.~9, no.~17, pp. 16\,168--16\,180, 2022.

\bibitem{anantrasirichai2021fast}
N.~Anantrasirichai, M.~Geravand, D.~Braendler, and D.~R. Bull, ``Fast depth estimation for view synthesis,'' in \emph{2020 28th European signal processing conference (EUSIPCO)}.\hskip 1em plus 0.5em minus 0.4em\relax IEEE, 2021, pp. 575--579.

\bibitem{rudolph2022lightweight}
M.~Rudolph, Y.~Dawoud, R.~G{\"u}ldenring, L.~Nalpantidis, and V.~Belagiannis, ``Lightweight monocular depth estimation through guided decoding,'' in \emph{2022 International Conference on Robotics and Automation (ICRA)}.\hskip 1em plus 0.5em minus 0.4em\relax IEEE, 2022, pp. 2344--2350.

\bibitem{zhou2017unsupervised}
T.~Zhou, M.~Brown, N.~Snavely, and D.~G. Lowe, ``Unsupervised learning of depth and ego-motion from video,'' in \emph{Proceedings of the IEEE conference on computer vision and pattern recognition}, 2017, pp. 1851--1858.

\bibitem{bian2019unsupervised}
J.~Bian, Z.~Li, N.~Wang, H.~Zhan, C.~Shen, M.-M. Cheng, and I.~Reid, ``Unsupervised scale-consistent depth and ego-motion learning from monocular video,'' \emph{Advances in neural information processing systems}, vol.~32, 2019.

\bibitem{bochkovskii2024depth}
A.~Bochkovskii, A.~Delaunoy, H.~Germain, M.~Santos, Y.~Zhou, S.~R. Richter, and V.~Koltun, ``Depth pro: Sharp monocular metric depth in less than a second,'' \emph{arXiv preprint arXiv:2410.02073}, 2024.

\bibitem{pham2024sharpdepth}
D.-H. Pham, T.~Do, P.~Nguyen, B.-S. Hua, K.~Nguyen, and R.~Nguyen, ``Sharpdepth: Sharpening metric depth predictions using diffusion distillation,'' \emph{arXiv preprint arXiv:2411.18229}, 2024.

\bibitem{ramamonjisoa2019sharpnet}
M.~Ramamonjisoa and V.~Lepetit, ``Sharpnet: Fast and accurate recovery of occluding contours in monocular depth estimation,'' in \emph{Proceedings of the IEEE/CVF International Conference on Computer Vision Workshops}, 2019, pp. 0--0.

\bibitem{yu2023udepth}
B.~Yu, J.~Wu, and M.~J. Islam, ``Udepth: Fast monocular depth estimation for visually-guided underwater robots,'' in \emph{2023 IEEE International Conference on Robotics and Automation (ICRA)}.\hskip 1em plus 0.5em minus 0.4em\relax IEEE, 2023, pp. 3116--3123.

\bibitem{poggi2022real}
M.~Poggi, F.~Tosi, F.~Aleotti, and S.~Mattoccia, ``Real-time self-supervised monocular depth estimation without gpu,'' \emph{IEEE Transactions on Intelligent Transportation Systems}, vol.~23, no.~10, pp. 17\,342--17\,353, 2022.

\bibitem{mungoli2023adaptive}
N.~Mungoli, ``Adaptive ensemble learning: Boosting model performance through intelligent feature fusion in deep neural networks,'' \emph{arXiv preprint arXiv:2304.02653}, 2023.

\bibitem{xia2021multi}
Y.~Xia, K.~Chen, and Y.~Yang, ``Multi-label classification with weighted classifier selection and stacked ensemble,'' \emph{Information Sciences}, vol. 557, pp. 421--442, 2021.

\bibitem{lee2022mpvit}
Y.~Lee, J.~Kim, J.~Willette, and S.~J. Hwang, ``Mpvit: Multi-path vision transformer for dense prediction,'' in \emph{Proceedings of the IEEE/CVF conference on computer vision and pattern recognition}, 2022, pp. 7287--7296.

\bibitem{garg2016unsupervised}
R.~Garg, V.~K. Bg, G.~Carneiro, and I.~Reid, ``Unsupervised cnn for single view depth estimation: Geometry to the rescue,'' in \emph{Computer Vision--ECCV 2016: 14th European Conference, Amsterdam, The Netherlands, October 11-14, 2016, Proceedings, Part VIII 14}.\hskip 1em plus 0.5em minus 0.4em\relax Springer, 2016, pp. 740--756.

\bibitem{godard2019digging}
C.~Godard, O.~Mac~Aodha, M.~Firman, and G.~J. Brostow, ``Digging into self-supervised monocular depth estimation,'' in \emph{Proceedings of the IEEE/CVF international conference on computer vision}, 2019, pp. 3828--3838.

\bibitem{xie2016deep3d}
J.~Xie, R.~Girshick, and A.~Farhadi, ``Deep3d: Fully automatic 2d-to-3d video conversion with deep convolutional neural networks,'' in \emph{Computer Vision--ECCV 2016: 14th European Conference, Amsterdam, The Netherlands, October 11--14, 2016, Proceedings, Part IV 14}.\hskip 1em plus 0.5em minus 0.4em\relax Springer, 2016, pp. 842--857.

\bibitem{godard2017unsupervised}
C.~Godard, O.~Mac~Aodha, and G.~J. Brostow, ``Unsupervised monocular depth estimation with left-right consistency,'' in \emph{Proceedings of the IEEE conference on computer vision and pattern recognition}, 2017, pp. 270--279.

\bibitem{zhan2018unsupervised}
H.~Zhan, R.~Garg, C.~S. Weerasekera, K.~Li, H.~Agarwal, and I.~Reid, ``Unsupervised learning of monocular depth estimation and visual odometry with deep feature reconstruction,'' in \emph{Proceedings of the IEEE conference on computer vision and pattern recognition}, 2018, pp. 340--349.

\bibitem{zhou2019unsupervised}
W.~Zhou, E.~Zhou, G.~Liu, L.~Lin, and A.~Lumsdaine, ``Unsupervised monocular depth estimation from light field image,'' \emph{IEEE Transactions on Image Processing}, vol.~29, pp. 1606--1617, 2019.

\bibitem{sun2023sc}
L.~Sun, J.-W. Bian, H.~Zhan, W.~Yin, I.~Reid, and C.~Shen, ``Sc-depthv3: Robust self-supervised monocular depth estimation for dynamic scenes,'' \emph{IEEE Transactions on Pattern Analysis and Machine Intelligence}, 2023.

\bibitem{tan2023deep}
J.~Tan, M.~Gao, T.~Duan, and X.~Gao, ``A deep joint network for monocular depth estimation based on pseudo-depth supervision,'' \emph{Mathematics}, vol.~11, no.~22, p. 4645, 2023.

\bibitem{wang2021knowledge}
Y.~Wang, X.~Li, M.~Shi, K.~Xian, and Z.~Cao, ``Knowledge distillation for fast and accurate monocular depth estimation on mobile devices,'' in \emph{Proceedings of the IEEE/CVF Conference on Computer Vision and Pattern Recognition}, 2021, pp. 2457--2465.

\bibitem{spek2018cream}
A.~Spek, T.~Dharmasiri, and T.~Drummond, ``Cream: Condensed real-time models for depth prediction using convolutional neural networks,'' in \emph{2018 IEEE/RSJ International Conference on Intelligent Robots and Systems (IROS)}.\hskip 1em plus 0.5em minus 0.4em\relax IEEE, 2018, pp. 540--547.

\bibitem{wofk2019fastdepth}
D.~Wofk, F.~Ma, T.-J. Yang, S.~Karaman, and V.~Sze, ``Fastdepth: Fast monocular depth estimation on embedded systems,'' in \emph{2019 International Conference on Robotics and Automation (ICRA)}.\hskip 1em plus 0.5em minus 0.4em\relax IEEE, 2019, pp. 6101--6108.

\bibitem{zhang2023lite}
N.~Zhang, F.~Nex, G.~Vosselman, and N.~Kerle, ``Lite-mono: A lightweight cnn and transformer architecture for self-supervised monocular depth estimation,'' in \emph{Proceedings of the IEEE/CVF Conference on Computer Vision and Pattern Recognition}, 2023, pp. 18\,537--18\,546.

\bibitem{zhou2021r}
Z.~Zhou, X.~Fan, P.~Shi, and Y.~Xin, ``R-msfm: Recurrent multi-scale feature modulation for monocular depth estimating,'' in \emph{Proceedings of the IEEE/CVF international conference on computer vision}, 2021, pp. 12\,777--12\,786.

\bibitem{xue2021boundary}
F.~Xue, J.~Cao, Y.~Zhou, F.~Sheng, Y.~Wang, and A.~Ming, ``Boundary-induced and scene-aggregated network for monocular depth prediction,'' \emph{Pattern Recognition}, vol. 115, p. 107901, 2021.

\bibitem{jung2021fine}
H.~Jung, E.~Park, and S.~Yoo, ``Fine-grained semantics-aware representation enhancement for self-supervised monocular depth estimation,'' in \emph{Proceedings of the IEEE/CVF International Conference on Computer Vision}, 2021, pp. 12\,642--12\,652.

\bibitem{ochs2019sdnet}
M.~Ochs, A.~Kretz, and R.~Mester, ``Sdnet: Semantically guided depth estimation network,'' in \emph{Pattern Recognition: 41st DAGM German Conference, DAGM GCPR 2019, Dortmund, Germany, September 10--13, 2019, Proceedings 41}.\hskip 1em plus 0.5em minus 0.4em\relax Springer, 2019, pp. 288--302.

\bibitem{miclea2023dynamic}
V.-C. Miclea and S.~Nedevschi, ``Dynamic semantically guided monocular depth estimation for uav environment perception,'' \emph{IEEE Transactions on Geoscience and Remote Sensing}, 2023.

\bibitem{lyu2021hr}
X.~Lyu, L.~Liu, M.~Wang, X.~Kong, L.~Liu, Y.~Liu, X.~Chen, and Y.~Yuan, ``Hr-depth: High resolution self-supervised monocular depth estimation,'' in \emph{Proceedings of the AAAI conference on artificial intelligence}, vol.~35, no.~3, 2021, pp. 2294--2301.

\bibitem{ke2024repurposing}
B.~Ke, A.~Obukhov, S.~Huang, N.~Metzger, R.~C. Daudt, and K.~Schindler, ``Repurposing diffusion-based image generators for monocular depth estimation,'' in \emph{Proceedings of the IEEE/CVF Conference on Computer Vision and Pattern Recognition}, 2024, pp. 9492--9502.

\bibitem{gui2024depthfm}
M.~Gui, J.~Schusterbauer, U.~Prestel, P.~Ma, D.~Kotovenko, O.~Grebenkova, S.~A. Baumann, V.~T. Hu, and B.~Ommer, ``Depthfm: Fast monocular depth estimation with flow matching,'' \emph{arXiv preprint arXiv:2403.13788}, 2024.

\bibitem{saxena2024surprising}
S.~Saxena, C.~Herrmann, J.~Hur, A.~Kar, M.~Norouzi, D.~Sun, and D.~J. Fleet, ``The surprising effectiveness of diffusion models for optical flow and monocular depth estimation,'' \emph{Advances in Neural Information Processing Systems}, vol.~36, 2024.

\bibitem{zhao2023gasmono}
C.~Zhao, M.~Poggi, F.~Tosi, L.~Zhou, Q.~Sun, Y.~Tang, and S.~Mattoccia, ``Gasmono: Geometry-aided self-supervised monocular depth estimation for indoor scenes,'' in \emph{Proceedings of the IEEE/CVF International Conference on Computer Vision}, 2023, pp. 16\,209--16\,220.

\bibitem{howard2017mobilenets}
A.~G. Howard, M.~Zhu, B.~Chen, D.~Kalenichenko, W.~Wang, T.~Weyand, M.~Andreetto, and H.~Adam, ``Mobilenets: Efficient convolutional neural networks for mobile vision applications,'' \emph{arXiv preprint arXiv:1704.04861}, 2017.

\bibitem{yang2024depth}
L.~Yang, B.~Kang, Z.~Huang, X.~Xu, J.~Feng, and H.~Zhao, ``Depth anything: Unleashing the power of large-scale unlabeled data,'' in \emph{Proceedings of the IEEE/CVF Conference on Computer Vision and Pattern Recognition}, 2024, pp. 10\,371--10\,381.

\bibitem{wang2004image}
Z.~Wang, A.~C. Bovik, H.~R. Sheikh, and E.~P. Simoncelli, ``Image quality assessment: from error visibility to structural similarity,'' \emph{IEEE transactions on image processing}, vol.~13, no.~4, pp. 600--612, 2004.

\bibitem{cheng2024gam}
A.~Cheng, Z.~Yang, H.~Zhu, and K.~Mao, ``Gam-depth: Self-supervised indoor depth estimation leveraging a gradient-aware mask and semantic constraints,'' \emph{arXiv preprint arXiv:2402.14354}, 2024.

\bibitem{sun2024dynamo}
Y.~Sun and B.~Hariharan, ``Dynamo-depth: fixing unsupervised depth estimation for dynamical scenes,'' \emph{Advances in Neural Information Processing Systems}, vol.~36, 2024.

\bibitem{wang2024weatherdepth}
J.~Wang, C.~Lin, L.~Nie, S.~Huang, Y.~Zhao, X.~Pan, and R.~Ai, ``Weatherdepth: Curriculum contrastive learning for self-supervised depth estimation under adverse weather conditions,'' in \emph{2024 IEEE International Conference on Robotics and Automation (ICRA)}.\hskip 1em plus 0.5em minus 0.4em\relax IEEE, 2024, pp. 4976--4982.

\bibitem{silberman2012indoor}
N.~Silberman, D.~Hoiem, P.~Kohli, and R.~Fergus, ``Indoor segmentation and support inference from rgbd images,'' in \emph{Computer Vision--ECCV 2012: 12th European Conference on Computer Vision, Florence, Italy, October 7-13, 2012, Proceedings, Part V 12}.\hskip 1em plus 0.5em minus 0.4em\relax Springer, 2012, pp. 746--760.

\bibitem{geiger2013vision}
A.~Geiger, P.~Lenz, C.~Stiller, and R.~Urtasun, ``Vision meets robotics: The kitti dataset,'' \emph{The International Journal of Robotics Research}, vol.~32, no.~11, pp. 1231--1237, 2013.

\bibitem{koch2018evaluation}
T.~Koch, L.~Liebel, F.~Fraundorfer, and M.~Korner, ``Evaluation of cnn-based single-image depth estimation methods,'' in \emph{Proceedings of the European Conference on Computer Vision (ECCV) Workshops}, 2018, pp. 0--0.

\bibitem{eigen2014depth}
D.~Eigen, C.~Puhrsch, and R.~Fergus, ``Depth map prediction from a single image using a multi-scale deep network,'' \emph{Advances in neural information processing systems}, vol.~27, 2014.

\end{thebibliography}
 

\end{document}